\documentclass[conference]{IEEEtran}
\IEEEoverridecommandlockouts
\usepackage{colortbl} 
\usepackage{cite}
\usepackage{multirow}
\usepackage{authblk}
\usepackage{amsmath,amssymb,amsfonts}
\usepackage{algorithmic}
\usepackage{graphicx}
\usepackage{textcomp}
\usepackage{subcaption}
\usepackage{xcolor}
\usepackage{soul}
\usepackage[colorinlistoftodos]{todonotes}
\usepackage{fontawesome5} 
\usepackage{pifont}

\def\BibTeX{{\rm B\kern-.05em{\sc i\kern-.025em b}\kern-.08em
    T\kern-.1667em\lower.7ex\hbox{E}\kern-.125emX}}

\begin{document}
\renewcommand\Authands{,}

\newcommand{\affiliations}[2]{%
  \vspace{-\bigskipamount}  
  \begin{center}
    \Affilfont
    \textsuperscript{1}#1,
    \textsuperscript{2}#2
  \end{center}
}

\title{Leveraging Registers in Vision Transformers for Robust Adaptation\\
}
%

\author[1]{Srikar Yellapragada}
\author[2]{Kowshik Thopalli}
\author[2]{Vivek Narayanaswamy}
\author[2]{Wesam Sakla} 
\author[2]{\\Yang Liu}
\author[2]{Yamen Mubarka}
\author[1]{Dimitris Samaras}
\author[2]{ Jayaraman J. Thiagarajan}

\affil[1]{Stony Brook University}
\affil[2]{Lawrence Livermore National Laboratory}

\maketitle

\begin{abstract}

Vision Transformers (ViTs) have shown success across a variety of tasks due to their ability to capture global image representations. Recent studies have identified the existence of high-norm tokens in ViTs, which can interfere with unsupervised object discovery. To address this, the use of "registers" which are additional tokens that isolate high norm patch tokens while capturing global image-level information has been proposed. While registers have been studied extensively for object discovery, their generalization properties particularly in out-of-distribution (OOD) scenarios, remains underexplored. In this paper, we examine the utility of register token embeddings in providing additional features for improving generalization and anomaly rejection. To that end, we propose a simple method that combines the special CLS token embedding commonly employed in ViTs with the average-pooled register embeddings to create feature representations which are subsequently used for training a downstream classifier. We find that this enhances OOD generalization and anomaly rejection, while maintaining in-distribution (ID) performance. Extensive experiments across multiple ViT backbones trained with and without registers reveal consistent improvements of 2-4\% in top-1 OOD accuracy and a 2-3\% reduction in false positive rates for anomaly detection. Importantly, these gains are achieved without additional computational overhead.

\end{abstract}

\section{Introduction}

\begin{figure}[t]
    \centering
    \includegraphics[width=1\linewidth]{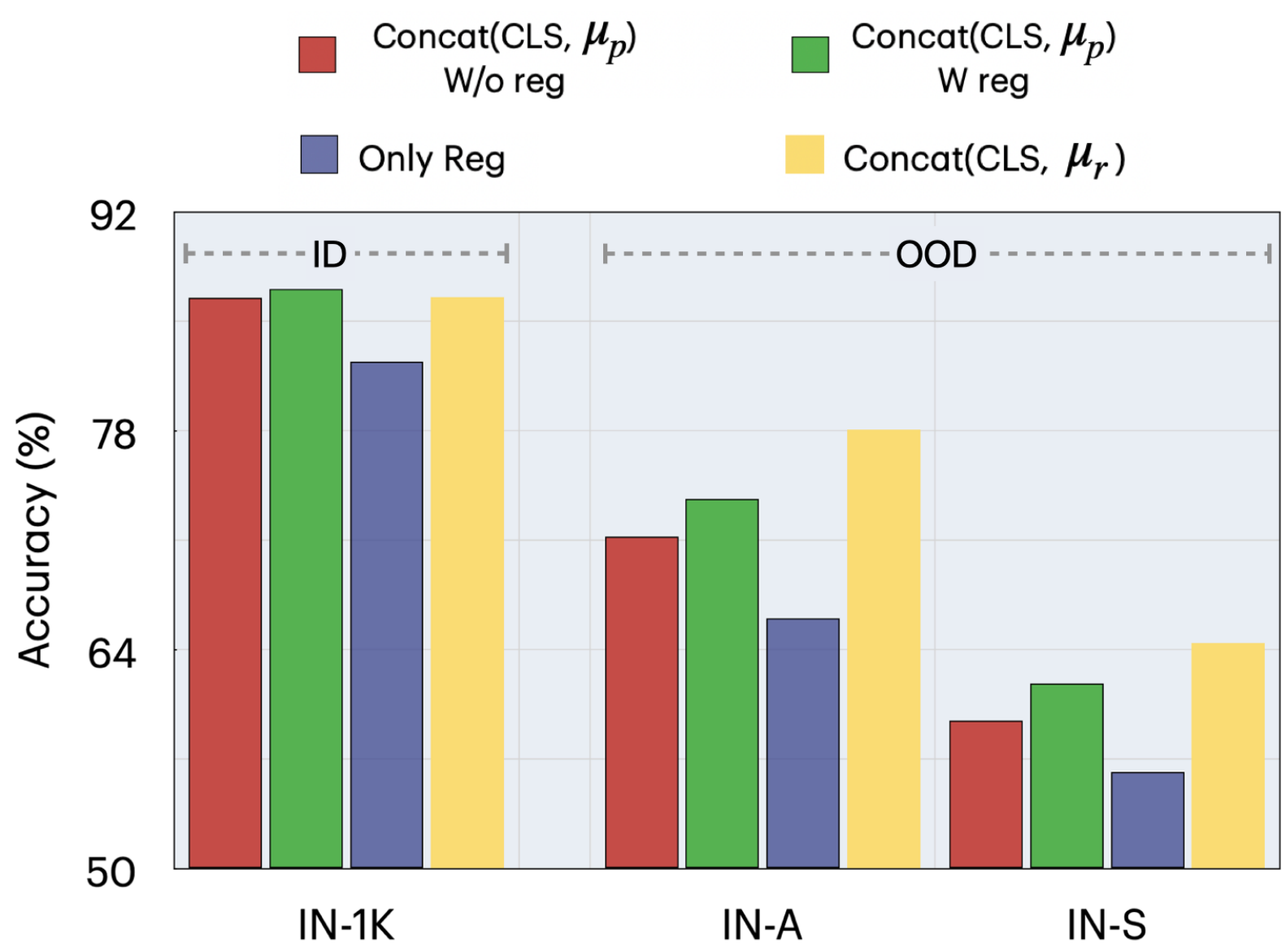}
    \caption{\textbf{Impact of token embedding choices on linear probing on frozen Dino-V2 ViT-G backbones. } Each color indicates the token embeddings chosen for optimizing a linear classifier on ImageNet (IN)-1K along with the protocol adopted for pre-training the backbone (w/o registers or w registers). Here, \texttt{[CLS]} represents the classification token, $\mu_P$ and $\mu_R$ represents the mean patch and register token embeddings respectively. While it is common to utilize \texttt{[CLS]} and $\mu_P$ to train the classifier, we obtain improved generalization if the backbone is trained with registers (red vs green). Training a classifier naively with register tokens results in drop in generalization. However, our approach (yellow) maintains ID accuracy while providing substantial gains in OOD (ImageNet-Adversarial, ImageNet-Sketch) accuracies.}
    \label{fig:motivation}
\end{figure}

Vision Transformers (ViTs)~\cite{dosovitskiy2021an}, which adapt the transformer architecture from natural language processing~\cite{vaswani2017attention}, have demonstrated exceptional performance across a range of vision tasks including classification and object detection~\cite{touvron2022deit, dettmers2022gptint, peebles2023scalable}. At a high level, ViTs process images by dividing them into fixed-size patches, each treated as a token, with a special \texttt{[CLS]} token appended to aggregate information across tokens for downstream tasks.

In a recent study on the behavior of large scale ViTs,  Darcet \textit{et al.}~\cite{darcet2024vision} observed that patch tokens associated with low-informative regions (e.g., background) tend to have significantly higher $\ell_2$ norm values compared to other tokens. While this tendency does not impact their utility in image-level prediction tasks, it increases the risk of compromising the performance in dense prediction tasks (e.g., object discovery). To circumvent this behavior, the authors explored the use of ``registers", which are additional tokens appended to the input sequence during training. Interestingly, these registers were found not only to isolate the behavior of the high norm patch tokens but also contain global image-level information. This was validated when a linear classifier was trained on register token embeddings extracted from in-distribution (ID) data, demonstrating enhanced classification performance.

While the utility of registers in achieving improving object discovery has been well studied, their efficacy in downstream tasks such as classification remains under explored, which is the primary focus of our paper. To thoroughly assess the role of registers in these tasks, it is critical to evaluate their generalization capabilities under a variety of out-of-distribution (OOD) scenarios.  As shown in Figure~\ref{fig:motivation}, naively training a linear classifier with register token embeddings results in a non-trivial drop in generalization performance on OOD datasets implying that the registers by themselves do not capture all the information required to achieve robustness.

Consequently, in this paper, we examine the utility of  register token embeddings, which are shown to contain global image-level information, as auxiliary features that supplement the \texttt{[CLS]} token embedding. We believe that such a combination will yield in enriched representations for robust adaptation. This hypothesis aligns with the findings of Zhang \textit{et al.}~\cite{zhang2023learning}, where seemingly redundant features with respect to in-distribution data provided significant benefits in OOD scenarios when combined together.

\begin{figure*}[!t]
    \centering
    \includegraphics[width=\linewidth]{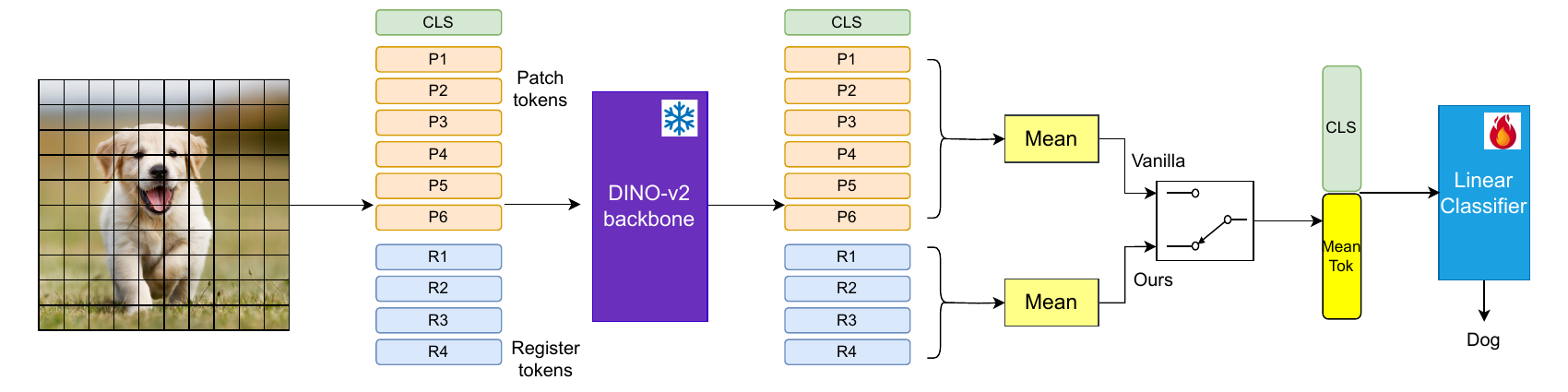}
    \caption{\textbf{Overview of our proposed method}: For large-scale vision transformer backbones (e.g., DINO-v2) pre-trained with ``registers", we find that concatenating registers (Mean (R$_1$, R$_2$, R$_3$, R$_4$) along with \texttt{[CLS]} is critical for obtaining rich features that enable robust adaptation. In particular, we train a linear classifier on these concatenated features and observe improved generalization and anomaly rejection capabilities.}
    \label{fig:method}
\end{figure*}

To that end, we train a linear classifier by combining the representations from the \texttt{[CLS]} token with the average-pooled register token embeddings $\mu_R$. We compare our approach with the standard practice of combining \texttt{[CLS]} token with average-pooled patch tokens $\mu_P$ \cite{oquab2023dinov2}. Strikingly, our method significantly improves both OOD generalization, while maintaining ID performance, as shown in Fig \ref{fig:motivation}. Moreover, our approach also boosts anomaly rejection performance, compared to baselines. Our comprehensive experiments across multiple ViT backbones reveal a consistent improvement of \textbf{2-4\%} in top-1 accuracy for OOD generalization and an average reduction of \textbf{2-3\%} in the false positive rate for anomaly detection. These findings support our hypothesis that the concatenation of registers with \texttt{[CLS]} yields more robust feature representations. Importantly, these gains are obtained without incurring any additional computational overhead.

In summary, our contributions are as follows:
\begin{itemize}
    \item We develop a novel method that leverages the register token embeddings as auxiliary features for building robust adaptation protocols.
    \item Through extensive experiments, we demonstrate consistent improvements in OOD generalization and anomaly detection across multiple ViT architectures.
    \item Our method achieves these gains with negligible computational overhead during training or inference, offering an efficient solution.
\end{itemize}
\section{Related Work}
\noindent\textbf{Large activations in transformers:} 
Understanding the behavior of transformers has been an important research topic in the last few years, owing to their success in various NLP and vision tasks. A common approach employed to achieve this is to analyze the activations and the features extracted by the model. Recent studies have uncovered the presence of outlier weights and large-magnitude activations in transformers, affecting both language \cite{heimersheim2023residual,kovaleva2021bert,zhao2023unveiling,timkey2021all} and vision models \cite{darcet2024vision,mambaregisters}. These phenomena manifest as concentrated attention weights in LLMs \cite{kovaleva2021bert,zhao2023unveiling} and in vision models~\cite{darcet2024vision}

\noindent\textbf{Implications and solutions for massive activations:} These large activations are found to pose challenges for model quantization \cite{dettmers2022gptint,he2024understanding} and image segmentation \cite{darcet2024vision}. Proposed remedies to address these challenges include appending ``register'' tokens to the input patch sequence \cite{darcet2024vision} or explicit attention biases \cite{sun2024massive} during training. While these approaches typically discard the additional tokens during inference, our work identifies that these tokens contain valuable auxiliary information to that captured by the \texttt{[CLS]} token. Therefore, we propose leveraging this information to obtain richer feature representations, enabling superior performance across a diverse range of tasks.

\section{Method}



\noindent\textbf{Preliminaries:}
Vision Transformers (ViTs) process images by dividing them into patches and viewing them as a sequence of tokens, where each token corresponds to a patch in the image. An additional token, the \texttt{[CLS]} token, is appended to the patch sequence to aggregate the global information from the image patches. The \texttt{[CLS]} embedding is often used as the global representation of the image in downstream tasks. 

Let $X \in \mathbb{R}^{H \times W \times 3}$ represent a $3$ channel RGB input image, where $H$ and $W$ are the height and width of the image, respectively. A ViT backbone, denoted by $F$, divides $X$ into $L$ patches, each of size $K \times K$, and projects them into $D$-dimensional embeddings. The \texttt{[CLS]} token $c \in \mathbb{R}^D$ is added to aggregate global information. The backbone produces:  

\begin{itemize}
    \item \texttt{[CLS]} token $c \in \mathbb{R}^D$
    \item Patch tokens $P = \{p_1 \ldots p_L \} \in \mathbb{R}^{L \times D} $
    \item Register tokens $R = \{r_1 \ldots r_M \} \in \mathbb{R}^{M \times D}$
\end{itemize}

where $D$ is the hidden dimensionality of the transformer, and $L$ and $M$ are the numbers of patch tokens and register tokens, respectively. 

In typical ViT models, downstream tasks use either the \texttt{[CLS]} token or the mean of patch tokens $\mu_P = \frac{1}{L} \sum_{j=1}^L p_j$ \cite{Zhai_2022_CVPR,li2024inverse}. In Dino-v2, the \texttt{[CLS]} token is concatenated with $\mu_p$, and fed to a linear classifier $h_{\theta}$ which is optimized with the following objective 

\begin{align}
\min_{\theta} \frac{1}{N} \sum_{i=1}^{N} \ell(h_{\theta}(f_i), y_i)
\label{eq:ERM}
\end{align}

where $N$ is the total number of samples, $f_i = [c_i; \mu_P^i]$ is the concatenated  \texttt{[CLS]} token \( c_i \) and the mean of the patch tokens \( \mu_P^i \) for the \(i^\text{th}\) sample, $h_{\theta}(f_i) = f_i^\top \theta$ denotes the linear probe with parameters \( \theta \), and \( \ell(\cdot, \cdot) \) is the cross-entropy loss computed between the prediction \( h_{\theta}(f_i) \) and the target \( y_i \). \\

\noindent\textbf{Proposed Approach:}
We propose to leverage the auxiliary information in register tokens to augment the information found in the \texttt{[CLS]} token, thereby enhancing out-of-distribution (OOD) detection performance. As previously stated, Darcet \textit{et al.} \cite{darcet2024vision} noted that register tokens capture global information similar to the \texttt{[CLS]} token, by training linear probes on register tokens without observing significant drop in ID performance. Furthermore, Zhang et al \cite{zhang2023learning} showed that although similar representations from models trained with different random seeds yield comparable ID performance, they can encode different information under OOD conditions. Crucially, combining these representations leads to richer feature sets that outperform either representation alone in OOD tasks.

\renewcommand{\arraystretch}{1.4}
\begin{table*}[ht]
\centering
\caption{\textbf{Evaluating ID and OOD generalization as well anomaly rejection performance across different ViT architectures.} We highlight the best performing method in every case with green. While the ID accuracies remain comparable to a baseline variant, we find that our proposed approach consistently outperforms the existing baselines on OOD generalization as well as anomaly rejection across a variety of benchmarks and architectures.}

\renewcommand{\arraystretch}{1.4}
\resizebox{1\linewidth}{!}{
\begin{tabular}{|cccccccccccccc|}
\hline
\rowcolor[gray]{0.92} 
\multicolumn{14}{|c|}{ViT-Giant}                                                                                                                                                                                                                              \\ \hline
\multicolumn{1}{|c|}{\multirow{2}{*}{Method}} & \multicolumn{1}{c|}{\multirow{2}{*}{ \shortstack{Dino-v2 \\ with registers}}} & \multicolumn{1}{c|}{\multirow{2}{*}{ID Acc}} & \multicolumn{3}{c|}{ImageNet OOD}                                     & \multicolumn{8}{c|}{Anomaly Rejection (FPR $\downarrow$ / AUROC $\uparrow$)}                                                                          \\ \cline{4-14} 
\multicolumn{1}{|c|}{}                        & \multicolumn{1}{c|}{}                           & \multicolumn{1}{c|}{}                         & In-A           & In-R           & \multicolumn{1}{c|}{In-S}           & \multicolumn{1}{c|}{Score} & DTD         & SVHN        & Places-365  & LSUN        & LSUN-R      & ISUN        & Mean        \\ \hline
\multicolumn{1}{|c|}{\texttt{CLS} ; $\mu_P$}       & \multicolumn{1}{c|}{\ding{53}}                          & \multicolumn{1}{c|}{86.5}                    & 73.38          & 77.63           & \multicolumn{1}{c|}{61.24}          & \multicolumn{1}{c|}{\multirow{3}{*}{MSP}}   & 48.81/85.32 & 22.68/95.71   &  { {\color[HTML]{0f9845}\textbf{59.65}}}/83.0  & 18.76/96.04 & 43.32/90.95  & 41.9/90.32 & 39.19/90.22 \\
\multicolumn{1}{|c|}{\texttt{CLS} ; $\mu_P$}  & \multicolumn{1}{c|}{\color[HTML]{237A1E}\faCheckSquare{ }}                          & \multicolumn{1}{c|}{\color[HTML]{0f9845}\textbf{87.1}}                    & 75.38          & 79.05          & \multicolumn{1}{c|}{62.90}           & \multicolumn{1}{c|}{}    & 46.42/86.16  & 27.16/94.57 & 58.58/{{\color[HTML]{0f9845}\textbf{83.09}}} & 16.3/96.5 & 40.41/91.14  & 38.27/91.25 & 37.86/90.45 \\
\multicolumn{1}{|c|}{\texttt{CLS} ; $\mu_R$}  & \multicolumn{1}{c|}{\color[HTML]{237A1E}\faCheckSquare{ } }                          & \multicolumn{1}{c|}{86.57}   &  {\color[HTML]{0f9845}\textbf{78.09}} &  {\color[HTML]{0f9845}\textbf{80.51}} & \multicolumn{1}{c|}{ {\color[HTML]{0f9845}\textbf{64.43}}} & \multicolumn{1}{c|}{}    &  {\color[HTML]{0f9845}\textbf{45.53}}/ {\color[HTML]{0f9845}\textbf{86.07}} &  {\color[HTML]{0f9845}\textbf{16.71}}/ {\color[HTML]{0f9845}\textbf{96.71}} & {58.88}/82.79 &  {\color[HTML]{0f9845}\textbf{11.78}}/{\color[HTML]{0f9845}\textbf{97.48}} &  {\color[HTML]{0f9845}\textbf{37.06}}/{\color[HTML]{0f9845}\textbf{91.59}} &  {\color[HTML]{0f9845}\textbf{33.76}}/{\color[HTML]{0f9845}\textbf{92.22}} &  {\color[HTML]{0f9845}\textbf{33.95}}/{\color[HTML]{0f9845}\textbf{91.14}} \\\hline
\multicolumn{1}{|c|}{\texttt{CLS} ; $\mu_P$}       & \multicolumn{1}{c|}{\ding{53}}                          & \multicolumn{1}{c|}{-}                        & -              & -              & \multicolumn{1}{c|}{-}              & \multicolumn{1}{c|}{\multirow{3}{*}{Energy}} & 31.83/{{\color[HTML]{0f9845}\textbf{92.12}}} & 10.11/97.66 & 43.04/{{\color[HTML]{0f9845}\textbf{89.72}}} & 6.34/98.60  & 30.30/94.92 & 26.73/94.74 & 24.73/94.63 \\ 
\multicolumn{1}{|c|}{\texttt{CLS} ; $\mu_P$}  & \multicolumn{1}{c|}{\color[HTML]{237A1E}\faCheckSquare{ }}                          & \multicolumn{1}{c|}{-}                        & -              & -              & \multicolumn{1}{c|}{-}              & \multicolumn{1}{c|}{} & {{\color[HTML]{0f9845}\textbf{30.48}}}/91.45 & 6.81/98.52 & 43.17/88.84 & 5.03/98.85  & 21.77/{\color[HTML]{0f9845}\textbf{95.78}} & 20.25/95.70 & 21.25/{\color[HTML]{0f9845}\textbf{94.86}}\\ 
\multicolumn{1}{|c|}{\texttt{CLS} ; $\mu_R$}  & \multicolumn{1}{c|}{\color[HTML]{237A1E}\faCheckSquare{ }}                          & \multicolumn{1}{c|}{-}                        & -              & -              & \multicolumn{1}{c|}{-}              & \multicolumn{1}{c|}{} & 32.41/90.64 & {\color[HTML]{0f9845}\textbf{4.96}}/{\color[HTML]{0f9845}\textbf{98.92}}  & {\color[HTML]{0f9845}\textbf{45.5}}/{\color[HTML]{0f9845}\textbf{87.64}}  & {\color[HTML]{0f9845}\textbf{3.85}}/{\color[HTML]{0f9845}\textbf{99.06}}  & {\color[HTML]{0f9845}\textbf{19.95}}/95.76 & {\color[HTML]{0f9845}\textbf{18.48}}/{\color[HTML]{0f9845}\textbf{95.95}} & {\color[HTML]{0f9845}\textbf{20.86}}/94.66\\ \hline
\end{tabular}
}

\vspace{0.5em} 
\renewcommand{\arraystretch}{1.4}
\resizebox{1\linewidth}{!}{
\begin{tabular}{|cccccccccccccc|}
\hline
\rowcolor[gray]{0.92}
\multicolumn{14}{|c|}{ViT-Large}                                                                                                                                                                                                                     \\ \hline
\multicolumn{1}{|c|}{\multirow{2}{*}{Method}} & \multicolumn{1}{c|}{\multirow{2}{*}{ \shortstack{Dino-v2 \\ with registers}} } & \multicolumn{1}{c|}{\multirow{2}{*}{ID Acc}} & \multicolumn{3}{c|}{ImageNet OOD}                                    & \multicolumn{8}{c|}{Anomaly Rejection (FPR $\downarrow$ / AUROC $\uparrow$)}                                                                          \\ \cline{4-14} 
\multicolumn{1}{|c|}{}                        & \multicolumn{1}{c|}{}                           & \multicolumn{1}{c|}{}                             & In-A           & In-R          & \multicolumn{1}{c|}{In-S}           & \multicolumn{1}{c|}{Score} & DTD         & SVHN        & Places-365  & LSUN        & LSUN-R      & ISUN        & Mean        \\ \hline
\multicolumn{1}{|c|}{\texttt{CLS} ; $\mu_P$}                & \multicolumn{1}{c|}{\ding{53}}                          & \multicolumn{1}{c|}{86.3}                            & 71.24          & 74.52          & \multicolumn{1}{c|}{59.46}          & \multicolumn{1}{c|}{\multirow{3}{*}{MSP}} & 51.4/84.59 & 27.2/94.6 & 59.54/83.24 & 20.63/95.49 & 40.81/91.61 & 40.46/90.9    & 40.1/90.07 \\
\multicolumn{1}{|c|}{\texttt{CLS} ; $\mu_P$}           & \multicolumn{1}{c|}{\color[HTML]{237A1E}\faCheckSquare{ }}                          & \multicolumn{1}{c|}{\color[HTML]{0f9845}\textbf{86.7}}                            & 73.65          & 76.31         & \multicolumn{1}{c|}{61.86}          & \multicolumn{1}{c|}{}                     &48.12/85.67 & 22.44/95.42  & 58.13/83.54 & 17.36/96.32 & 37.35/92.39 & 37.49/91.48 & 36.82/90.81  \\
\multicolumn{1}{|c|}{\texttt{CLS} ; $\mu_R$}           & \multicolumn{1}{c|}{\color[HTML]{237A1E}\faCheckSquare{ }}                          & \multicolumn{1}{c|} {85.82}                            & {\color[HTML]{0f9845}\textbf{75.33}} & {\color[HTML]{0f9845}\textbf{79.2}} & \multicolumn{1}{c|}{\color[HTML]{0f9845}{\textbf{62.82}}} & \multicolumn{1}{c|}{}                     & {\color[HTML]{0f9845}\textbf{43.71}}/{\color[HTML]{0f9845}\textbf{86.25}} & {\color[HTML]{0f9845}\textbf{15.93}}/{\color[HTML]{0f9845}\textbf{96.76}} & {\color[HTML]{0f9845}\textbf{55.88}}/{\color[HTML]{0f9845}\textbf{83.35}} & {\color[HTML]{0f9845}\textbf{12.65}}/{\color[HTML]{0f9845}\textbf{97.12}} & {\color[HTML]{0f9845}\textbf{33.49}}/{\color[HTML]{0f9845}\textbf{92.38}} & {\color[HTML]{0f9845}\textbf{31.35}}/{\color[HTML]{0f9845}\textbf{92.77}} & {\color[HTML]{0f9845}\textbf{32.17}}/{\color[HTML]{0f9845}\textbf{91.44}}
 \\ \hline
\multicolumn{1}{|c|}{\texttt{CLS} ; $\mu_P$}                & \multicolumn{1}{c|}{\ding{53}}                          & \multicolumn{1}{c|}{-}                            & \textbf{-}     & -             & \multicolumn{1}{c|}{-}              & \multicolumn{1}{c|}{\multirow{3}{*}{Energy}} & 47.27/87.26 & 48.95/93.03 & 55.77/86.15 & 17.8/96.13 & 49.99/91.78 & 42.85/91.91 & 43.77/91.04  \\ 
\multicolumn{1}{|c|}{\texttt{CLS} ; $\mu_P$}           & \multicolumn{1}{c|}{\color[HTML]{237A1E}\faCheckSquare{ }}                          & \multicolumn{1}{c|}{-}                            & -              & -             & \multicolumn{1}{c|}{-}              & \multicolumn{1}{c|}{}   & 35.83/90.51  & 10.35/97.73 & 45.21/{\color[HTML]{0f9845}\textbf{89.05}} & 8.08/98.18 & 24.83/95.76 & 24.56/95.21 & 24.81/94.41 \\ 
\multicolumn{1}{|c|}{\texttt{CLS} ; $\mu_R$}           & \multicolumn{1}{c|}{\color[HTML]{237A1E}\faCheckSquare{ }}                          & \multicolumn{1}{c|}{-}                            & -              & -             & \multicolumn{1}{c|}{-}              & \multicolumn{1}{c|}{}   & {\color[HTML]{0f9845}\textbf{34.06}}/{\color[HTML]{0f9845}\textbf{90.87}} & {\color[HTML]{0f9845}\textbf{7.05}}/{\color[HTML]{0f9845}\textbf{98.55}} & {\color[HTML]{0f9845}\textbf{43.44}}/88.79 & {\color[HTML]{0f9845}\textbf{6.86}}/{\color[HTML]{0f9845}\textbf{98.5}} & {\color[HTML]{0f9845}\textbf{19.24}}/{\color[HTML]{0f9845}\textbf{96.21}} & {\color[HTML]{0f9845}\textbf{18.87}}/{\color[HTML]{0f9845}\textbf{96.15}} & {\color[HTML]{0f9845}\textbf{21.59}}/{\color[HTML]{0f9845}\textbf{94.84}}
\\ \hline
\end{tabular}

}

\vspace{0.5em}
\renewcommand{\arraystretch}{1.4}
\resizebox{1\linewidth}{!}{
\begin{tabular}{|cccccccccccccc|}
\hline
\rowcolor[gray]{0.92}
\multicolumn{14}{|c|}{ViT-Base}                                                                                                                                                                                                                       \\ \hline
\multicolumn{1}{|c|}{\multirow{2}{*}{Method}} & \multicolumn{1}{c|}{\multirow{2}{*}{ \shortstack{Dino-v2 \\ with registers}}} & \multicolumn{1}{c|}{\multirow{2}{*}{ID Acc}} & \multicolumn{3}{c|}{ImageNet OOD}                                     & \multicolumn{8}{c|}{Anomaly Rejection (FPR $\downarrow$ / AUROC $\uparrow$)}                                                                          \\ \cline{4-14} 
\multicolumn{1}{|c|}{}                        & \multicolumn{1}{c|}{}                           & \multicolumn{1}{c|}{}                             & In-A           & In-R           & \multicolumn{1}{c|}{In-S}           & \multicolumn{1}{c|}{Score} & DTD         & SVHN        & Places-365  & LSUN        & LSUN-R      & ISUN        & Mean        \\ \hline
\multicolumn{1}{|c|}{\texttt{CLS} ; $\mu_P$}                & \multicolumn{1}{c|}{\ding{53}}                          & \multicolumn{1}{c|}{\color[HTML]{0f9845}\textbf{84.5} }                           & 55.54 & 63.9         & \multicolumn{1}{c|}{50.92}          & \multicolumn{1}{c|}{\multirow{3}{*}{MSP}}   & 55.66/83.95 & {\color[HTML]{0f9845}\textbf{18.35}}/{\color[HTML]{0f9845}\textbf{96.44}}  & 64.65/81.76 & 21.94/95.42 & 39.51/91.57 & 39.9/91.24 & 40.01/90.06 \\
\multicolumn{1}{|c|}{\texttt{CLS} ; $\mu_P$}           & \multicolumn{1}{c|}{\color[HTML]{237A1E}\faCheckSquare{ }}                          & \multicolumn{1}{c|}{84.21}                           & 54.56          & 65.02          & \multicolumn{1}{c|}{53.61}          & \multicolumn{1}{c|}{}    & 53.12/84.45  & 20.31/96.27 & 62.87/82.34 & 16.84/96.31 & 33.83/92.98 & 35.38/92.24 & 37.06/90.76  \\
\multicolumn{1}{|c|}{\texttt{CLS} ; $\mu_R$}           & \multicolumn{1}{c|}{\color[HTML]{237A1E}\faCheckSquare{ }}                          & \multicolumn{1}{c|}{83.84}                            & {\color[HTML]{0f9845}\textbf{57.87}}          &{\color[HTML]{0f9845}\textbf{68.75}} & \multicolumn{1}{c|}{\color[HTML]{0f9845}\textbf{55.14}} & \multicolumn{1}{c|}{}    & {\color[HTML]{0f9845}\textbf{48.09}}/{\color[HTML]{0f9845}\textbf{85.42}}& 19.83/95.8 & {\color[HTML]{0f9845}\textbf{59.6}}/{\color[HTML]{0f9845}\textbf{82.48}} & {\color[HTML]{0f9845}\textbf{13.55}}/{\color[HTML]{0f9845}\textbf{97.09}} & {\color[HTML]{0f9845}\textbf{30.24}}/{\color[HTML]{0f9845}\textbf{93.66}} & {\color[HTML]{0f9845}\textbf{29.56}}/{\color[HTML]{0f9845}\textbf{93.36}} & {\color[HTML]{0f9845}\textbf{33.48}}/{\color[HTML]{0f9845}\textbf{91.3}}
 \\\hline
\multicolumn{1}{|c|}{\texttt{CLS} ; $\mu_P$}                & \multicolumn{1}{c|}{\ding{53}}                          & \multicolumn{1}{c|}{-}                            & \textbf{-}     & -              & \multicolumn{1}{c|}{-}              & \multicolumn{1}{c|}{\multirow{3}{*}{Energy}} & 60.09/85.47 & 47.68/93.64 & 74.61/82.53 & 33.55/94.50 & 70.7/88.61 & 64.44/89.28 & 58.51/89.01 \\
\multicolumn{1}{|c|}{\texttt{CLS} ; $\mu_P$}           & \multicolumn{1}{c|}{\color[HTML]{237A1E}\faCheckSquare{ }}                          & \multicolumn{1}{c|}{-}                            & \textbf{-}     & \textbf{-}     & \multicolumn{1}{c|}{\textbf{-}}     & \multicolumn{1}{c|}{} & 42.32/89.58 & 18.22/96.34 & 52.38/87.91 & 8.73/98.06     & 24.85/95.82 & 24.03/95.44 & 28.59/93.86 \\
\multicolumn{1}{|c|}{\texttt{CLS} ; $\mu_R$}           & \multicolumn{1}{c|}{\color[HTML]{237A1E}\faCheckSquare{ }}                          & \multicolumn{1}{c|}{-}                            & -              & -              & \multicolumn{1}{c|}{-}              & \multicolumn{1}{c|}{} & {\color[HTML]{0f9845}\textbf{37.54}}/{\color[HTML]{0f9845}\textbf{90.71}} & {\color[HTML]{0f9845}\textbf{12.02}}/{\color[HTML]{0f9845}\textbf{97.46}} & {\color[HTML]{0f9845}\textbf{46.38}}/{\color[HTML]{0f9845}\textbf{88.83}} & {\color[HTML]{0f9845}\textbf{7.07}}/{\color[HTML]{0f9845}\textbf{98.57}} & {\color[HTML]{0f9845}\textbf{17.91}}/{\color[HTML]{0f9845}\textbf{96.97}} & {\color[HTML]{0f9845}\textbf{16.87}}/{\color[HTML]{0f9845}\textbf{96.74}} & {\color[HTML]{0f9845}\textbf{22.96}}/{\color[HTML]{0f9845}\textbf{94.88}}
 \\ \hline
\end{tabular}
}
\label{tab:results}
\end{table*}

Building on this insight, we concatenate the \texttt{[CLS]} token with the mean of register tokens, $\mu_R = \frac{1}{M} \sum_{k=1}^M r_k$, as opposed to discarding the register tokens as described in previous works  \cite{darcet2024vision, hoptimus0}.  This combined representation exploits the auxiliary information captured by the register tokens, allowing us to create a richer feature representation. The concatenated representation is then fed to the linear classifier $h_{\theta}$ which is optimized with the same objective as presented in Eq. \ref{eq:ERM} except that the $f_i$ is now given by  $f_i = [c_i; \mu_R^i]$. Figure. \ref{fig:method} provides an overview of the proposed approach. 




Note our method does not introduce additional computational overhead because the ViT backbone is already trained and kept frozen during the experiments. We only need to train a linear layer, however, just like the baseline approach that concatenates the \texttt{[CLS]} token with the patch token mean. Since both methods involve concatenating two vectors ($[\texttt{CLS}; \mu_R]$ vs $[\texttt{CLS}; \mu_P]$), the linear layer remains the same size. Thus, given a backbone already trained with register tokens, the only step required is training this linear layer, making our approach computationally efficient.



\section{Experiments}
\noindent{\textbf{Datasets}}. We evaluate our proposed approach on the following datasets: \underline{(i) Training/In-Distribution (ID) data}: We train the Linear Classifiers (LC) on ImageNet-1K~\cite{russakovsky2015imagenet} for all experiments. ImageNet-1K is a large-scale imaging benchmark comprising $1.3$ million training images and $50,000$ validation images across $1000$ diverse categories. \underline{(ii) Out-of-Distribution (OOD) generalization benchmarks}: (a) ImageNet-A (In-A)~\cite{hendrycks2021natural} is a dataset comprising $7,500$ adversarially filtered images across a $200$-classes subset of ImageNet-1K’s $1,000$ classes that were found to be challenging to current ImageNet classifiers, (b) ImageNet-R (In-R)~\cite{hendrycks2021many} contains $30,000$ images of different real-world renditions from $200$ classes of ImageNet; (c) ImageNet-S (In-S)~\cite{wang2019learning} contains $50$ black and white sketch images for each class in ImageNet for a total of $50,000$. \underline{(iii) Anomaly rejection benchmarks}: Following standard practice~\cite{energyood}, we consider (a) Texture (DTD), (b) SVHN, (c) Places 365, (d) LSUN, (e) LSUN (Resized) and (f) iSUN datasets. It must be noted that, every image in these datasets is appropriately resized and passed as input into the networks.  

\noindent{\textbf{Setup}}. We utilize the open-source\footnote{https://github.com/facebookresearch/dinov2} Dino-v2 ViT backbones trained with registers~\cite{darcet2024vision} for the experiments involving register tokens, and Dino-v2 ViT backbone trained without registers for other experiments.  We evaluate three variants: ViT Giant, Large, and Base. In all cases, we only train the last linear layer of the network while keeping the backbone frozen. We train the linear layers with the SGD optimizer for $10,000$ iterations, using random-resized-crop data augmentation. The training process was conducted on $16$ NVIDIA V$100$ GPUs, with a batch size of $256$ per GPU and learning rate of $0.01$.

\noindent{\textbf{Metrics}}. We report the Top@1 accuracy to measure OOD generalization performance. For anomaly rejection, we utilize Maximum Softmax Probability (MSP)~\cite{hendrycks2017a} and energy~\cite{energyood} scoring functions and report the FPR@TPR$95$ and AUROC~\cite{energyood} for performance evaluation.  

\noindent{\textbf{Baselines}}. We compare our proposed approach with the following methods to understand the impact of choosing different tokens for training the LC. (i) Concatenation of the [\texttt{CLS}] token and the patch token mean (\texttt{CLS}; $\mu_p$ W/o reg) when the model is trained without registers, as in \cite{oquab2023dinov2}, and (ii) the same concatenation when the model is trained with registers (\texttt{CLS}; $\mu_p$ with reg). To ensure consistent feature scales during concatenation, we use features extracted after the final LayerNorm from the Dinov2 backbones.

\subsection{OOD Generalization}
As shown in Table \ref{tab:results}, our experiments on OOD generalization demonstrate the effectiveness of combining the [\texttt{CLS}] token with the mean of Register tokens (\texttt{CLS}; $\mu_R$). Across all three ViT architectures (Giant, Large and Base), our approach consistently outperforms both the baselines. For ViT-Giant, our method achieves the highest accuracy on all three ImageNet OOD generalization benchmarks. This represents improvements of 2.71\%, 1.46\% and 1.53\% respectively over the best performing baseline of (\texttt{CLS}; $\mu_p$ with reg). Similar trends are observed for ViT-Large and ViT-Base. These results suggest that the combination of by viewing register tokens as auxiliary features and combining it with \texttt{CLS} embeddings yield substantially superior capture more features in terms of generalization and robustness. 
\subsection{Anomaly Rejection}
Our approach demonstrates even significant improvements for the task of anomaly rejection and consistently outperforms other ablations. For instance, with ViT-Base, FPR@TPR95 drops dramatically on SVHN (from 18.22 \% to 12.02\%) and LSUN-R (from 24.85 \% to 17.91 \%). On an average for ViT-Large, our method reduces the mean FPR by 22.18 and 3.22  percentage points compared to the baselines respectively. These improvements hold across all ViT architectures, demonstrating the robustness of our approach. Combining [\texttt{CLS}] and Register tokens enhances the model's ability to detect anomalies by capturing auxiliary information that helps distinguish ID from OOD samples. These findings indicate potential for effective real-world anomaly rejection.

\section{Conclusion}
In this work, we explored the utility of register token embeddings as auxilliary features that supplement the \texttt{[CLS]} token in Vision Transformers (ViTs) to improve the feature richness for out-of-distribution (OOD) generalization and anomaly rejection. While previous approaches discard register tokens after training, we demonstrated that these tokens contain valuable information that augments the global features captured by the \texttt{[CLS]} token. By concatenating the \texttt{[CLS]} token with the mean of register tokens, we produced richer representations for the linear classifier to perform better on OOD tasks across multiple ViT architectures pre-trained with or without registers. Our approach consistently outperformed baseline methods in both generalization and anomaly rejection, achieving notable improvements in top-1 accuracy and reductions in false positive rates. Importantly, these gains were achieved without introducing additional computational overhead. These findings suggest that register tokens are not redundant but instead play a crucial role in enhancing the robustness and adaptability of ViTs. 

\section*{Acknowledgment}
This work was performed under the auspices of the U.S. Department of Energy by the Lawrence Livermore National Laboratory under Contract No. DE-AC52-07NA27344, Lawrence Livermore National Security, LLC. This work is supported by LDRD projects 22-ERD-006 and 24-FS-002. LLNL-CONF-870596.

\bibliographystyle{unsrt}
\bibliography{main}

\vspace{12pt}

\end{document}